\documentclass[10pt,twocolumn,letterpaper]{article}
\usepackage{cvpr}      
\usepackage{multirow}
\definecolor{cvprblue}{rgb}{0.21,0.49,0.74}
\usepackage[accsupp]{axessibility}  
\usepackage[pagebackref,breaklinks,colorlinks,allcolors=cvprblue]{hyperref}
\def\paperID{10835} 
\def\confName{CVPR}
\def\confYear{2025}

\title{ASAP: Advancing Semantic Alignment Promotes Multi-Modal Manipulation \\Detecting and Grounding}

\author{
    \textbf{Zhenxing Zhang\textsuperscript{1}}, 
    \textbf{Yaxiong Wang\textsuperscript{1}\thanks{Corresponding authors: yaxiong.wang15@gmail.com}}, 
    \textbf{Lechao Cheng\textsuperscript{1}},
    \textbf{Zhun Zhong\textsuperscript{1}},
    \textbf{Dan Guo\textsuperscript{1,2*}},
    \textbf{Meng Wang\textsuperscript{1,2}}\\
    \textsuperscript{1}School of Computer Science and Information Engineering, Hefei University of Technology, China\\
    \textsuperscript{2}Institute of Artificial Intelligence, Hefei Comprehensive National Science Center, China\quad 
}
\begin{document}
\maketitle
\begin{abstract}
\label{sec:intro}
We present ASAP, a new framework for detecting and grounding multi-modal media manipulation (DGM\textsuperscript{4}). 
Upon thorough examination, we observe that accurate fine-grained cross-modal semantic alignment between the image and text is vital for accurately manipulation detection and grounding. While existing DGM\textsuperscript{4} methods pay rare attention to the cross-modal alignment, hampering the accuracy of manipulation detecting to step further.  To remedy this issue, this work targets to advance the semantic alignment learning to promote this task.  Particularly, we utilize the off-the-shelf large models to construct paired image-text pairs, especially for the manipulated instances. Subsequently, a cross-modal alignment learning is performed to enhance the semantic alignment. Besides the explicit auxiliary clues, we further design  a  Manipulation-Guided Cross Attention (MGCA) to provide implicit guidance for augmenting the manipulation perceiving.  With the grounding truth available during training, MGCA encourages the model to concentrate more on manipulated components while downplaying normal ones, enhancing the model's ability to capture manipulations. Extensive experiments are conducted on the DGM\textsuperscript{4} dataset, the results demonstrate that our model can surpass the comparison method with a clear margin. Code will be released at \href{https://github.com/CriliasMiller/ASAP}{https://github.com/CriliasMiller/ASAP}.
\end{abstract}  
\section{Introduction}
\label{sec:intro}

In recent years, the field of artificial intelligence has witnessed exponential growth, particularly with the advent of transformer architectures~\cite{transformer,vit} and diffusion models~\cite{diffusion,DIT,relay_diffusion}. These sophisticated models have given rise to large generative models capable of producing highly realistic outputs across modalities such as images, text, and audio~\cite{GPT4,Gemini,MMCode,QWEN,GLM}. The quality of these synthetic creations is so high that they can often deceive human perception. Concurrently, models for content editing have also seen swift development. As these technologies progress, the methods for manipulating visual and textual content have become increasingly complex, presenting a substantial threat to public information security. Consequently, the problem of Detecting and Grounding Multi-Modal Media Manipulation (DGM\textsuperscript{4}) has been proposed~\cite{HAMMER}, and extensive efforts have been dedicated to this field recently~\cite{HAMMER++,VIKI,UFTFormer}.

\begin{figure}[t]
\centering
\includegraphics[width=0.48\textwidth]{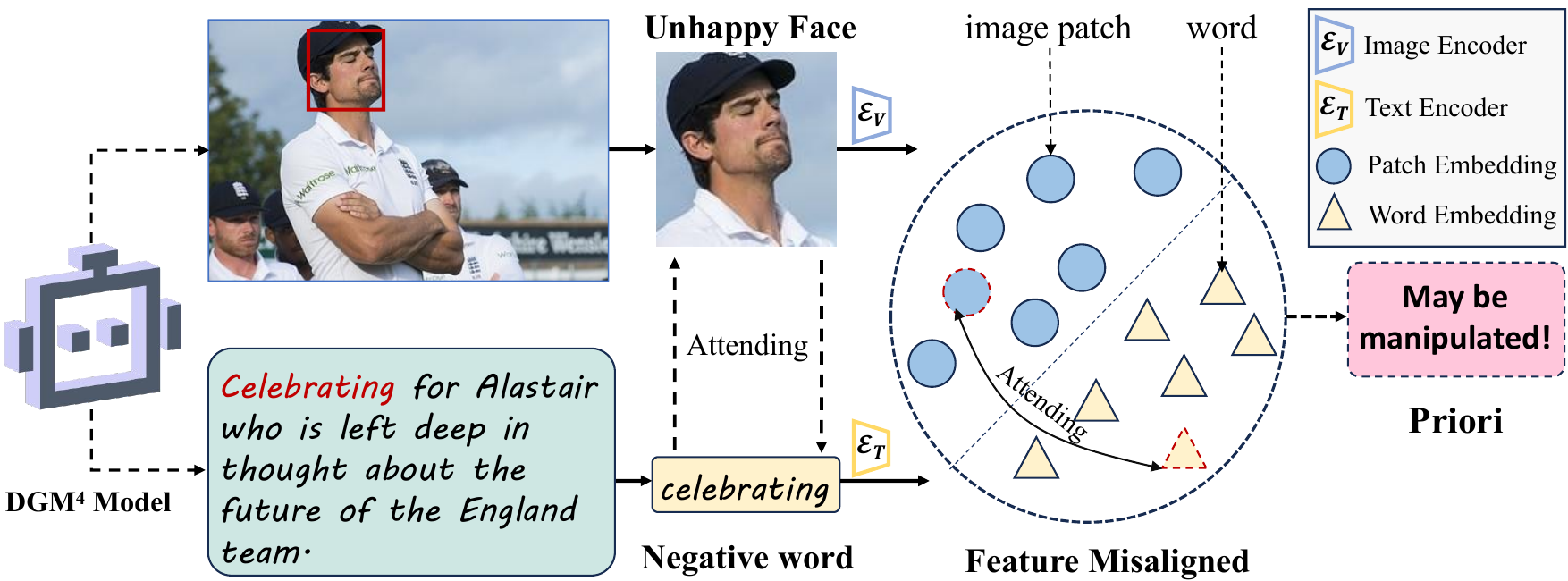} 
\caption{Fine-grained understanding of the multimodal media is one of keys for detecting the manipulated media. The capture of the misaligned components between the image and the text can effectively assist the DGM\textsuperscript{4} task.}
\label{teaser}
\end{figure}

DGM\textsuperscript{4} problem targets to detect whether the multimodal media is manipulated and locate the manipulated components like image regions and words in multimodal image-text inputs~\cite{HAMMER,HAMMER++}. The challenge of this task lies in the fine-grained understanding of the multimodal inputs, some attempts have been made to address this task~\cite{VIKI}. HAMMER~\cite{HAMMER} designs shallow manipulation reasoning module to locate the manipulated regions, and design a deep manipulation reasoning architecture for the overall recognition and grounding the text manipulations.VIKI~\cite{VIKI} detects disinformation in multimodal data by jointly grounding vision and language features through aligned embeddings. 
It optimizes these embeddings using metric learning and geometric distance minimization to ensure a compact hypothesis space. 
The model includes a knowledge interaction mechanism that integrates cross-modality information for improved multitask learning.

A notable truth for the DGM\textsuperscript{4} task is that if there is manipulation in a piece of media, the image and text are usually not aligned any more, and this statement can be validated by the setup of this task~\cite{HAMMER}. With this critical observation, we argue that one of the primary keys for DGM\textsuperscript{4} models is to effectively perceive the fine-grained image-text alignment in the media~\cite{pfan,sigir24,shuyu}. As depicted in Figure~\ref{teaser}, if the model can capture that the word "celebrating" does not align with the "disappointment" emotion on the human face, then the model can acquire the valuable priori knowledge that the media may be manipulated. Existing methods predominantly rely on annotations to directly pinpoint the manipulated components, but they often simply apply the alignment learning on the unchanged multimodal media, which is not sufficiently effective to assist the subsequent detecting and grounding.

Addressing the aforementioned issues, this paper centers on improving fine-grained alignment learning for the DGM\textsuperscript{4} task. Two key challenges arise: firstly, manipulated media's image and text are unpaired, precluding the use of conventional cross-modal alignment techniques such as contrastive learning~\cite{pfan,pfan1,sigir24}. In response, we propose leveraging off-the-shelf Multimodal Large Language Models (MLLMs) to generate descriptions for both manipulated and non-manipulated images~\cite{vlp}, followed by implementing contrastive learning between the generated captions and the respective images to fortify cross-modal alignment.

Another challenge to achieve the fine-grained alignment is the huge semantic gap between the social image and text. Unlike the general image-text pair that the text is formed by general words and the image often shows broad scenes, the description and the image in social media usually contains specific entities, which poses a significant challenges for the cross-modal alignment. For example, it is difficult for the model to align the image without the knowledge about a specific person or place mentioned in the text.  In response to this challenge, we treat off-the-shelf large language models as the knowledge base and query the explanatory text for the entity-specific description in multimodal media. Taking the explanatory text as the bridging clue, we subsequent perform the cross-modal contrastive learning on the explanatory text and the image to enhance the multimodal alignment, thereby facilitating the subsequent manipulation detecting and grounding.

In addition to providing explicit auxiliary cues, we introduce a Manipulation-Guided Cross Attention (MGCA) to prioritize attention on manipulated elements for improved detection and grounding. Given annotations accessible during training, MGCA first constructs a guidance mask and then employs a classification-guided constraint to adaptively modify the image-text attention matrix, directing the model to allocate heightened attention to manipulated regions while attenuating non-manipulated parts, thus simplifying subsequent detection and grounding tasks. With the above design, we finally develop our \textbf{A}dvancing \textbf{S}emantic \textbf{A}lignment framework to \textbf{P}romote (ASAP) the task of Detecting and Grounding Multi-Modal Manipulation. 
In summary, we highlight the contributions of this paper as follows:
    \begin{itemize}
        \item We disclose the significance of cross-modal alignment for DGM\textsuperscript{4} problem, which is paid rare attention by the previous methods. To remedy, a ASAP framework is proposed to enhance the semantic alignment to promote this task.
        \item To bolster cross-modal alignment, we devise a Large Model-assisted Alignment (LMA) approach that incorporates auxiliary text, encompassing captions and explanatory texts from large models, to enhance fine-grained semantic matching.
        \item A Manipulation-Guided Cross Attention (MGCA) is devised to facilitate the model's perception of manipulated components in the feature space. This is realized by allocating increased attention to manipulated elements within the attention matrix.
    \end{itemize}
\section{Relate Work}
\label{sec:relate}

\begin{figure*}[h]
\centering
\includegraphics[width=1\textwidth]{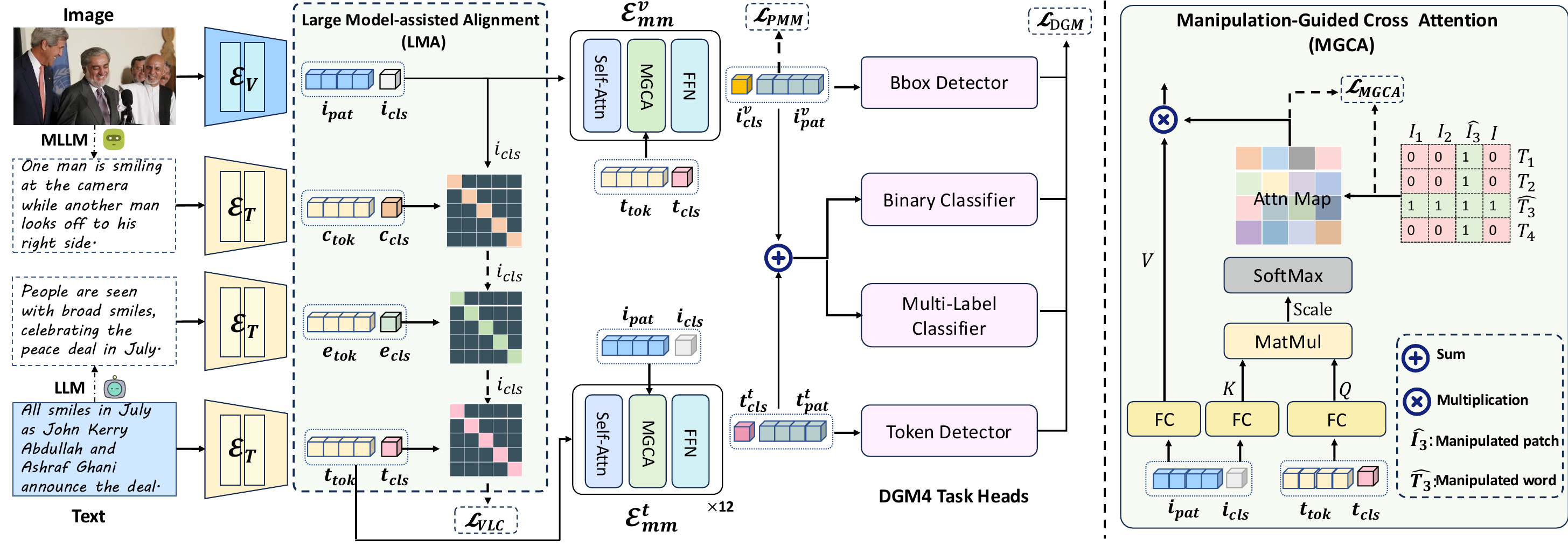} 
\caption{Illustration of our proposed ASAP framework. We employ a Multimodal Large Language Model (MLLM) to generate captions and a Large Language Model (LLM) to produce explanation texts for social media image-text pairs. These, along with the image, are encoded to obtain feature representations. Our Large Model-assisted Alignment (LMA) module enhances cross-modal alignment, followed by two Multimodal Encoders with Manipulation-Guided Cross Attention (MGCA) to integrate features for task-specific representations. One encoder is vision-biased for image grounding, and the other is text-biased for text grounding. The combined features from both encoders are used for media authenticity detection and manipulation identification. The network is optimized using DGM losses and objectives from LMA and MGCA.
}
\label{framework}
\end{figure*}
\noindent\textbf{Deepfake Detection.} The rapid advancement of generative models has accelerated the development of Deepfake detection technologies, which aim to uncover tampered information. Historically, detection methods were categorized into unimodal~\cite{Unimodel1Xue,Unimodel2,deepfake} and multimodal approaches. Among unimodal methods, GLFNet~\cite{Unimodel1Xue} combines physiological features with deep learning techniques to enhance detection performance. Recently, with the rise of multimodal large models and privacy concerns, multimodal Deepfake detection~\cite{MultilModel,MultilModel2} has gained prominence. The NewsCLIPpings dataset~\cite{NewsCLIP} has contributed to this field by providing mismatched image-text pairs. The DGM\textsuperscript{4} dataset further advances this area by covering diverse multimedia tampering scenarios and aligning more closely with practical applications~\cite{HAMMER}. Building on this, the HAMMER model captures fine-grained interactions between modalities and uses contrastive learning to assess image-text consistency.

\noindent\textbf{Large Models.} 
Large models have made significant strides in natural language processing (NLP) and multimodal tasks. Large language models (LLMs) like Mistral~\cite{Mistral7B} and LLaMA~\cite{llama} demonstrate impressive performance, with Mistral excelling in language understanding and generation, and LLaMA achieving high efficiency with fewer parameters. In the multimodal domain, models like BLIP2~\cite{Blip2} excel in image understanding and generation, while multimodal GPT-4~\cite{GPT4} integrates text and image processing for enhanced capabilities across tasks. Google's DeepMind Gato model~\cite{Gato} further pushes the boundaries with its general-purpose multimodal abilities. The growing adoption of large-scale models is driving innovation and providing novel solutions across various fields.
In this study, we leverage both LLMs and MLLMs to generate auxiliary texts, enhancing our model's performance in multimodal tasks.
\section{Methodology}
\label{method}
\noindent\textbf{Overview}. The architecture of our ASAP is depicted in Figure~\ref{framework}. In the training phase, we begin by leveraging a Multimodal Large Language Model (MLLM) to generate auxiliary caption and a Large Language Model (LLM) to produce explanation text for a given image-text pair from social media. Thereafter, the image and all associated texts are processed by separate image and text encoders to obtain their corresponding feature representations. Our Large Model-assisted Alignment (LMA) module then operates on these features to refine fine-grained cross-modal alignment. Successively, two Multimodal Encoders equipped with our Manipulation-Guided Cross Attention (MGCA) integrate the encoded visual and textual features for deriving task-centric representation. Concurrently, a vision-biased multimodal encoder focuses on capturing vision-dominant multimodal features to localize forged regions (image grounding), whereas a text-biased multimodal encoder encapsulates text-predominant features to pinpoint manipulated words (text grounding). The dual multimodal features are subsequently combined to facilitate both media authenticity detection and manipulation type identification. Ultimately, the entire network is optimized using the corresponding DGM\textsuperscript{4} losses and objectives stemming from our LMA and MGCA. 

{Note that all proposed modules only facilitate the alignment learning during training and are not deployed during inference.} Consequently, our ASAP framework imposes no additional computational load in the inference stage.

\subsection{Large Model-assisted Alignment.}
\noindent\textbf{Preliminary Text Generation.}
To enhance cross-model alignment, we first construct preliminary texts, including image captions and explanation texts for training ASAP.

\begin{figure}[t]
    \centering 
    \includegraphics[width=0.475\textwidth]{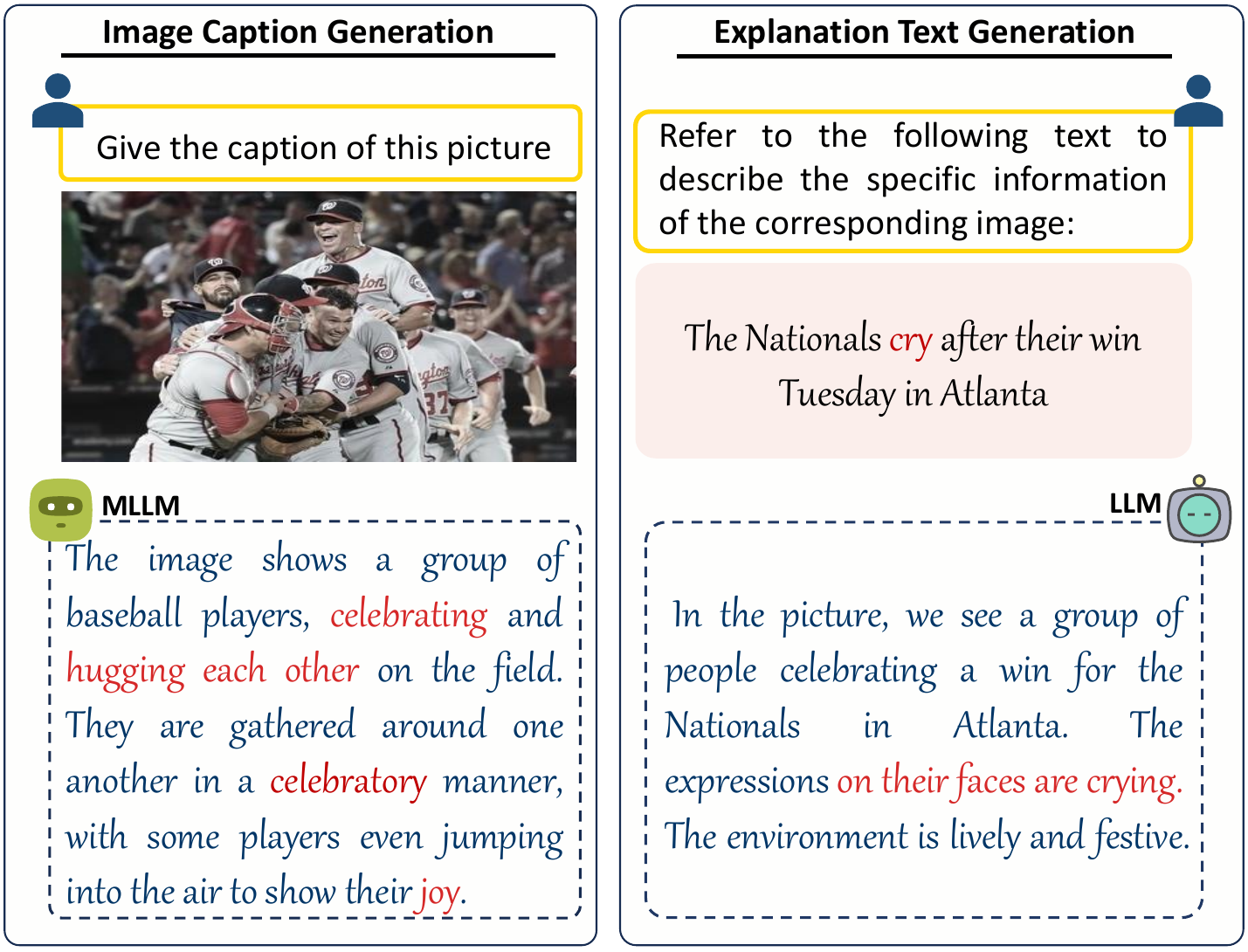} %
    \caption{Illustration of the generation of image caption (left) and explanation text (right).  The auxiliary texts can be effectively harvested via the off-the-shelf large models with the carefully crafted instructions. }
    \label{genexam} 
\end{figure}

To generate caption, we crafted an instruction: ``\texttt{Give the caption of this picture}'', and input this directive alongside the image into a pre-trained Multimodal Large Language Model (MLLM). Consequently, the detailed caption $C$ for the image is obtained. As depicted in the left subfigure of Figure~\ref{genexam}, the descriptive words generated, such as ``celebrating", ``hugging each other", ``jumping", and ``show their joy", correspond well with the image, contrasting with the mismatch of the word ``cry" in the textual description. 

In contrast to image captioning, which describes the image content by processing the image through MLLM, explanation text requires accurate ``imagining" of the image condition based on the textual description via a Large Language Model (LLM). To provide  a reliable guide for the LLM, we meticulously designed an instruction template: ``\texttt{Refer to the following text to describe the specific information of the corresponding image: [T]}", where $T$  is the textual input. The explanation text $E$ is generated by inputting this template into the LLM. As shown in the right panel of Figure~\ref{genexam}, the descriptive words derived from the text, such as ``celebrating", ``crying", and ``lively and festive", elucidate the specific image details. The explanation text provides an elaboration of the text with additional details. In cases of unaltered image and text, the explanation text can effectively correspond with the image, thus aiding in cross-modal alignment. For manipulated pairs, the explanation text offers a more detailed description of the discordant visual elements, as illustrated in Figure~\ref{genexam}. In the following, we would harness these properties of explanation text to bolster DGM\textsuperscript{4} task.

\noindent\textbf{Vision-Language Contrastive Learning.}
To enhance cross-modal alignment, we initially process the image $I$ through the visual encoder $\mathcal{E}_v$, yielding the feature set $\mathcal{E}_v(I) = \{ i_{\text{cls}}, i_{\text{pat}} \}$. Concurrently, the text $T$ is encoded by the text encoder $\mathcal{E}_t $ to extract text embeddings, denoted as $\mathcal{E}_t(T) = \{ t_{\text{cls}}, t_{\text{tok}} \}$. The subscript notation $ (\cdot)_{\text{cls}}$ indicates the classification token, while $(\cdot)_{\text{pat}}$ and $(\cdot)_{\text{tok}}$ signify the image patch and textual word tokens, respectively. Analogously, the auxiliary caption and explanation text are encoded with $\mathcal{E}_t $, yielding features $\mathcal{E}_t(C) = \{ c_{\text{cls}}, c_{\text{tok}} \}$ and $\mathcal{E}_t(E) = \{ e_{\text{cls}}, e_{\text{tok}} \}$, respectively.
 
For the generated caption, the image is always aligned with its caption. Consequently, we can utilize all image-caption pair to enhance the cross-modal alignment:  
\begin{equation}
\mathcal{L}_{i2c} =- \log \frac{\exp(s(i_{\text{cls}}, c_{\text{cls}}) / \tau)}{\sum_{\hat{c}_{\text{cls}}\in \mathcal{B}} \exp(s(i_{\text{cls}}, \hat{c}_{\text{cls}}) / \tau)},
\end{equation}
{where $\hat{c}_{\text{cls}}$ is the feature of unpaired caption with image in a mini-batch $\mathcal{B}$, $s(\cdot,\cdot)$ means to cosine similarity,  $\tau$ is a learnable parameter.}

Explanation text, however, originate from texts, cannot align with the image once the text or the image is manipulated. As a result, we only perform the contrastive learning among the image-explanation text pair where the image and the text are both unchanged. Formally, the constrain is formulated as: 

\begin{equation}
\mathcal{L}_{i2e} = -\mathbb{I}(\text{I},\text{T})\log \frac{\exp(s({i}_{\text{cls}}, e_{\text{cls}}) / \tau)}{\sum_{\hat{e}_{\text{cls}}\in\mathcal{B}} \exp(s({i}_{\text{cls}}, \hat{e}_{\text{cls}}) / \tau)},
\end{equation}
where $\mathbb{I}(\text{I},\text{T})=1$ if $I$ and $T$ are both not manipulated, 0 otherwise. $\hat{e}_{\text{cls}}$ is the feature of unpaired explanation text in the mini-batch.

Symmetrically, the caption-image and explanation-image contrastive loss, $\mathcal{L}_{c2i}$ and $\mathcal{L}_{e2i}$, can be formulated in a similar fashion. Finally, the overall objectives of our  vision-language contrastive learning is consolidated by further including the contrastive learning of the image and its associated text $T$:
\begin{equation}
    \label{vlcloss}
    \mathcal{L}_{\text{VLC}} = (\mathcal{L}_{i2t} + \mathcal{L}_{t2i})/2 + (\mathcal{L}_{i2c} + \mathcal{L}_{c2i} + \mathcal{L}_{i2e} + \mathcal{L}_{e2i})/4,
\end{equation}
where $\mathcal{L}_{i2t}$ and $\mathcal{L}_{t2i}$ are the contrastive loss of image-text and text-image.

\noindent\textbf{Explanation Text-Enhanced Manipulation Detecting.}
As shown in Figure~\ref{genexam}, explanation text provides another grained description for the image. Such a finer-grained linguistic text can on the one hand enhance the cross-modal alignment with contrastive learning. On the other hand, explanation text can also be treated as another counterpart of the original text to aid the manipulation detecting. In particular, the pair of image and explanation text from the manipulated media (either image or text is manipulated) is treated as the negative (fake) samples, while the ones from unchanged media (neither image nor text is manipulated) are taken as the positive (real) ones.

Formally, we first acquire the multimodal feature of image and explanation text via the multimodal encoder $\mathcal{E}_{mm}$ formed by self-attention and our MGCA. Subsequently, the multimodal feature 
then passes through the detecting head to get the prediction 
 $p(I, E)$ of the media is manipulated or not:
\begin{equation}
\label{detecthead}
\left\{\begin{aligned}
    &p(I, E) = \text{sigmoid}(\text{MLP}({\delta}F_{mm}^v+F_{mm}^t)),\\
    &F_{mm}^v=\mathcal{E}_{mm}^v(\mathcal{E}_v(I),\mathcal{E}_t(E)),\\
    &F_{mm}^t=\mathcal{E}_{mm}^t(\mathcal{E}_t(E),\mathcal{E}_v(I)),
\end{aligned}
\right.
\end{equation}
where $\mathcal{E}_{mm}^v$ is the vision-biased multimodal encoder with $\mathcal{E}_v(I)$ serving as the query, $\mathcal{E}_{mm}^t$ means text-biased encoder with $\mathcal{E}_t(E)$ as the query, $\delta$ is a learnable parameter. Next, the cross entropy loss is applied on the prediction:
\begin{equation}
\label{detectloss}
    \mathcal{L}_{\text{IED}} = -\mathbb{I}(\text{I},\text{T})\log p(I, E) - (1-\mathbb{I}(\text{I},\text{T}))\log(1-p(I, E)),
\end{equation}
Overall, our LMA includes two constrains:
\begin{equation}
    \mathcal{L}_{\text{LMA}} = \mathcal{L}_{\text{VLC}} + \mathcal{L}_{\text{IED}}.
\end{equation}

\subsection{Manipulation-Guided Cross Attention.}
\textbf{Motivation.} During our practice, we found that the commonly-used cross-attention in multimodal encoder  fails to allocate adequate attention to manipulated components. For example, within the text-biased multimodal encode $\mathcal{E}_{mm}^t$, the attention matrix is initially derived from interactions between textual tokens and image patches:
\begin{equation}
    \label{attmat}
    A = \text{softmax}(\frac{W_t\mathcal{E}_t(T)\times W_v\mathcal{E}_v(I)}{\sqrt{h}}),
\end{equation}
where $W_t$ and $W_v$ are two linear mapping matrix, $h$ is the dimension of the feature. The equation illustrates that the attention element \( A(i,j) \), indicative of the attention weight from the $i$-th word token to the $j$-th image patch, is proportional to the similarity between them. In the case of manipulated pairs, the manipulated word no longer aligns with certain image regions, resulting in diminished similarity. For instance, the word ``celebrating" does not correspond with an ``unhappy face" region, as exemplified in Figure~\ref{teaser}. This misalignment leads to a reduced attention value, contradicting the expected intuition that manipulated word should pay higher attention to the unmatched region shown in Figure~\ref{teaser}. As a result, the perceiving of the manipulations is impeded.

We introduce Manipulation-Guided Cross Attention (MGCA) to address this limitation. Leveraging the available word and region manipulation annotations from the training phase, we can formulate a guidance mask that directs the model's focus towards manipulated components.

\begin{equation}
G(i,j) = \begin{cases} 
1, & \text{if } I_i \text{ or } T_j \text{ is manipulated} \\
0, & \text{otherwise}
\end{cases}
\end{equation}
where $I_i$ and $T_j$ represent the $i$-th image patches and  $j$-th text token, respectively. We define an image patch $I_i$ is manipulated if it overlaps with the manipulation region.

With the manipulation guiding matrix $G$ established, we explicitly encourage the attention matrix for manipulated pairs to assign more attention to the manipulations. Mathematically, the cross attention is guided via following guidance constrain:

\begin{equation}
\mathcal{L}_{\text{MGCA}} = -\frac{1}{|A|}\sum_{i,j} \left[ G{(i,j)} \log \text{A}{(i,j)} \right],
\end{equation}
where $|A|$ means the number of elements in $A$.

\subsection{Training}

\begin{figure}[t]
    \centering 
    \includegraphics[width=0.46\textwidth]{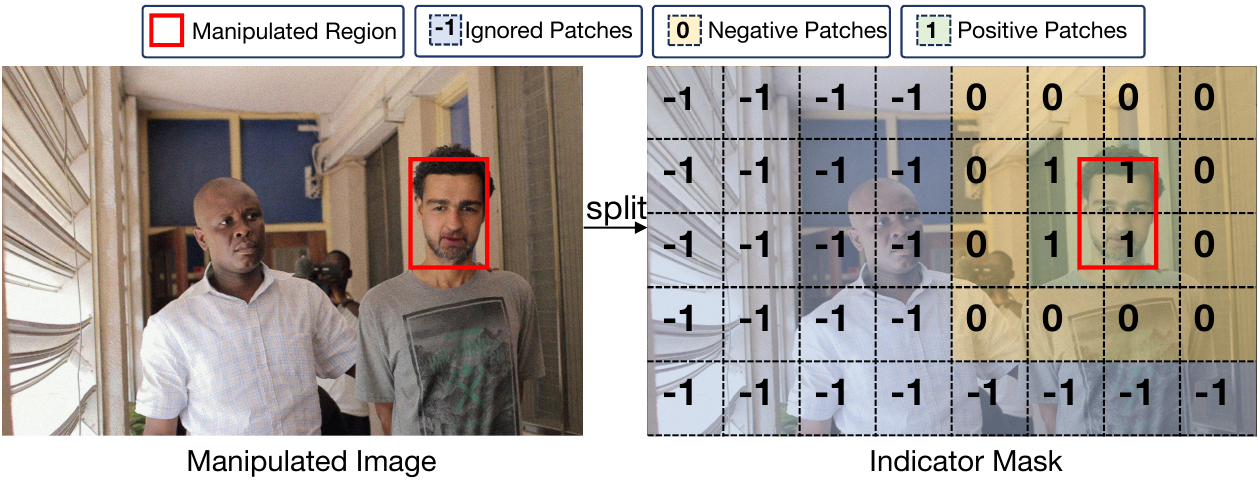} %
     \caption{Illustration of constructing the indicator mask. According to the manipulated region bounding box, the patches that ovelap with the box is taken the positive samples, while the adjacent patches to the positive patches are negative ones. The other patches are ignored. }
    \label{patchLoss} 
\end{figure}

\noindent\textbf{Patch Manipulation Modeling.} To ground the manipulated regions, the state-of-the-art methods usually use the global \texttt{[cls]} feature of multimodal feature from the multimodal encoder~\cite{HAMMER,VIKI}. However, we argue that the solely global representation lacks local patch contexts to accurately identify the regions. In response, we design a patch manipulation modeling that discriminates the manipulated patches before the final region bounding box prediction, thereby providing a prior for the final bounding box.   

Specifically, assume there are $N$ patches in the image. We initiate by creating a patch indicator $P \in \mathbb{R}^N$ to identify whether a patch is manipulated, based on the manipulation region annotations: $P_i = 1$ if the $i$-th patch intersects with a manipulated area, and 0 if it does not. Recognizing the spatial locality of images, patches distant from manipulated regions are trivially classified as negatives. Consequently, we propose a \emph{Hard Negative Patch Selection} (HNP) strategy: only adjacent patches to the manipulated ones are picked as the challenging negatives, as depicted in Figure~\ref{patchLoss}. The final patch indicator is thus defined accordingly:
\begin{equation}
{P}_i = \begin{cases}
1, & \text{if } I_i \text{ is manipulated}, \\
0, & \text{if } I_i \text{ is adjacent to manipulated patches}, \\
-1, & \text{otherwise (would be ignored)}.
\end{cases}    
\end{equation}
Subsequently, we harvest the multimodal patch feature  $i^v_\text{pat}$  from vision-biased multimodal encoder:
\begin{equation}
    \{i_\text{cls}^v, i_\text{pat}^v\} =  \mathcal{E}_{mm}^v(\mathcal{E}_v(I), \mathcal{E}_t(T)).
\end{equation}
Next, we discriminate the patches in hidden feature space via a binary classifier $\mathcal{C}$ and supervise the output with the constructed patch indicator:  
\begin{equation}
\begin{aligned}
\mathcal{L}_{\text{PMM}} &= -\frac{1}{N^{'}} \sum_{i_k\in i_{\text{pat}}^v, P_k\in\{0,1\}} [P_k \log \mathcal{C}(i_{k})\\
&+ (1 - {P}_k) \log(1 - \mathcal{C}(i_{k})],
\end{aligned}
\end{equation}
where $N^{'}$ is patch number marked as 1 or 0.

\begin{table*}[htbp]
\centering
\small
\setlength{\tabcolsep}{0.8mm}
\setlength{\extrarowheight}{0pt}{
\resizebox{.95\textwidth}{!}{
    \begin{tabular}{lccccccccccccc|c}
    \toprule[1.5pt]
    \rowcolor[gray]{.95}Tasks & \multirow{2}{*}{Reference} & \multicolumn{3}{c}{Binary Cls} & \multicolumn{3}{c}{Multi-Label} & \multicolumn{3}{c}{Image Grounding} & \multicolumn{3}{c}{Text Grounding} & \multirow{2}{*}{$\Delta$AVG} \\
    \cmidrule(r){3-5} \cmidrule(r){6-8}\cmidrule(r){9-11}\cmidrule(r){12-14}
    \rowcolor[gray]{.95}Methods & \ & AUC$\uparrow$ & EER$\downarrow$  & ACC$\uparrow$ & mAP$\uparrow$ & CF1$\uparrow$ & OF1$\uparrow$ & IoUmean$\uparrow$ & IoU50$\uparrow$ & IoU75$\uparrow$ & Precision$\uparrow$ & Recall$\uparrow$ & F1$\uparrow$ & \\
    \midrule
    CLIP~\cite{CLIP} & ICML21 & 83.22 & 24.61 & 76.40 & 66.00 & 59.52 & 62.31 & 49.51 & 50.03 & 38.79 & 58.12 & 22.11 & 32.03 & - \\
    VILT~\cite{VILT} & ICML21 & 85.16 & 22.88 & 78.38 & 72.37 & 66.14 & 66.00 & 59.32 & 65.18 & 48.10 & 66.48 & 49.88 & 57.00 & - \\
    HAMMER~\cite{HAMMER} & CVPR23 & 93.19 & 14.10 & 86.39 & 86.22 & 79.37 & 80.37 & 76.45 & 83.75 & 76.06 & 75.01 & 68.02 & 71.35 & 0 \\
    HAMMER++~\cite{HAMMER++} & TPAMI24 & 93.33 & 14.06  & 86.66 & 86.41 & 79.73 & 80.71 & 76.46 & 83.77 & 76.03 & 73.05 & 72.14 & 72.59 & +0.39 \\
    VIKI~\cite{VIKI} & IF24 & 93.51 & 13.87 & 86.67 & 86.58 & 81.07 & 80.10 & 76.51 & 83.95 & 75.77 & 77.79 & 66.06 & 72.44 & +0.37 \\
    UFAFormer~\cite{UFAFormer} & IJCV24 & 93.81 & 13.60 & 86.80 & 87.85 & 80.31 & 81.48 & \textbf{78.33} & \textbf{85.39} & \textbf{79.20} & 73.35 & 70.73 & 72.02 & +1.13 \\
    \textbf{ASAP(Ours)} & CVPR25 & \textbf{94.38} & \textbf{12.73} & \textbf{87.71} & \textbf{88.53} & \textbf{81.72} & \textbf{82.89} & 77.35 & 84.75 & 76.54 & \textbf{79.38} & \textbf{73.86} & \textbf{76.52} & +2.40 \\
    \bottomrule[1.5pt]
    \end{tabular}}
}
\caption{\textbf{Quantitative comparison} of our proposed ASAP method with the current state-of-the-art techniques across four key tasks: manipulation detection, manipulation type identification, image grounding, and text grounding. As evident from the comparison, ASAP achieves the top performance, showing a significant average improvement over the baseline HAMMER method in all sub-tasks. \textbf{$\Delta$AVG} indicates the improvement compared to HAMMER.}
\label{tab:comparison_HAMMER}
\end{table*}

\begin{table*}[htbp]
    \centering
    \small
    \setlength{\tabcolsep}{1.1mm}
    \setlength{\extrarowheight}{0pt}{
        \begin{tabular}{lcccccccccccc}
        \toprule[1.5pt]
        \rowcolor[gray]{.95}Tasks & \multicolumn{3}{c}{Binary Cls} &\multicolumn{3}{c} {Multi-Label} &  \multicolumn{3}{c}{Image Grounding} & \multicolumn{3}{c}{Text Grounding} \\
        \cmidrule(r){2-4} \cmidrule(r){5-7}\cmidrule(r){8-10}\cmidrule(r){11-13}
        \rowcolor[gray]{.95}ASAP Components & AUC$\uparrow$ & EER$\downarrow$  & ACC$\uparrow$ & mAP$\uparrow$ & CF1$\uparrow$ & OF1$\uparrow$ & IoUmean$\uparrow$ & IoU50$\uparrow$ & IoU75$\uparrow$ & Precision$\uparrow$ & Recall$\uparrow$ & F1$\uparrow$ \\
        \midrule
        Baseline & 93.16 & 14.13 & 86.23 & 86.23 & 79.59 & 80.54 & 76.49 & 83.82 & 75.97 & 75.25 & 68.21 & 71.83 \\
        +LMA & 94.28 &12.86 & 87.53 & 88.10 & 81.71 & 82.61 & 75.90 & 83.27 & 74.98 & 78.59 & \textbf{74.10} & 76.28 \\
        +LMA+MGCA & \textbf{94.40} &12.81 & 87.59 & 88.37 & \textbf{81.77} & 82.78 & 76.95 & 84.30 & 76.31 & 78.90 & 74.08 & 76.40 \\
        +LMA+MGCA+PMM & 94.38 & \textbf{12.73} & \textbf{87.71} & \textbf{88.53} & 81.72 & \textbf{82.89} & \textbf{77.35} & \textbf{84.75} & \textbf{76.54} & \textbf{79.38} & 73.86 & \textbf{76.52} \\
        \bottomrule[1.5pt]
        \end{tabular}
    }
    \caption{\textbf{Ablation study} of the components in our ASAP, including Large Model-aided Alignment (LMA), Manipulation-Guided Cross-Attention (MGCA), and the Patch Manipulation Modeling.}
    \label{tab:ablation}
\end{table*}

\begin{table*}[htbp]
\centering
\small
    \setlength{\tabcolsep}{1.1mm}
    \setlength{\extrarowheight}{0pt}{
        \begin{tabular}{lcccccccccccc}
        \toprule[1.5pt]
        \rowcolor[gray]{.95}Tasks & \multicolumn{3}{c}{Binary Cls} &\multicolumn{3}{c} {Multi-Label} &  \multicolumn{3}{c}{Image Grounding} & \multicolumn{3}{c}{Text Grounding} \\
        \cmidrule(r){2-4} \cmidrule(r){5-7}\cmidrule(r){8-10}\cmidrule(r){11-13}
        \rowcolor[gray]{.95}Auxiliary Text & AUC$\uparrow$ & EER$\downarrow$  & ACC$\uparrow$ & mAP$\uparrow$ & CF1$\uparrow$ & OF1$\uparrow$ & IoUmean$\uparrow$ & IoU50$\uparrow$ & IoU75$\uparrow$ & Precision$\uparrow$ & Recall$\uparrow$ & F1$\uparrow$ \\
        \midrule
        ASAP w/ \textbf{C} & 94.18 &12.81 & 87.60 & 87.89 & 81.60 & 82.73 & 76.93 & 84.29 & 76.30 & \textbf{79.98} & 73.28 & 76.48 \\
        ASAP w/ \textbf{E} & 94.22 &12.89 & 87.64 & 88.20 & 81.42 & 82.80 & 77.02 & 84.55 & 76.38 & 79.20 & 73.45 & 76.46 \\
        ASAP w/ \textbf{C \& E} &  \textbf{94.38} & \textbf{12.73} & \textbf{87.71} & \textbf{88.53} & \textbf{81.72} & \textbf{82.89} & \textbf{77.35} & \textbf{84.75} & \textbf{76.54} & 79.38 & \textbf{73.86} & \textbf{76.52} \\
        \bottomrule[1.5pt]
        \end{tabular}
    }
\caption{\textbf{Discussion the effectiveness of auxiliary caption(C) and explanation text(E).} Utilizing either the caption or the explanation text alone can individually enhance performance; however, incorporating both yields the optimal results.}
\label{tab:caption}
\end{table*}

\begin{table*}[htbp]
\centering
\small
    \setlength{\tabcolsep}{1.1mm}
    \setlength{\extrarowheight}{0pt}{
        \begin{tabular}{ccccccccccccc}
        \toprule[1.5pt]
        \rowcolor[gray]{.95}Tasks & \multicolumn{3}{c}{Binary Cls} &\multicolumn{3}{c} {Multi-Label} &  \multicolumn{3}{c}{Image Grounding} & \multicolumn{3}{c}{Text Grounding} \\
        \cmidrule(r){2-4} \cmidrule(r){5-7}\cmidrule(r){8-10}\cmidrule(r){11-13}
        \rowcolor[gray]{.95}Hard Negatives Or Not & AUC$\uparrow$ & EER$\downarrow$  & ACC$\uparrow$ & mAP$\uparrow$ & CF1$\uparrow$ & OF1$\uparrow$ & IoUmean$\uparrow$ & IoU50$\uparrow$ & IoU75$\uparrow$ & Precision$\uparrow$ & Recall$\uparrow$ & F1$\uparrow$ \\
        \midrule
        {ASAP w/o HNP} & 94.35 &12.82 & 87.55 & \textbf{88.59} & \textbf{81.73} & 82.67 & 76.92 & 84.24 & 76.15 & 79.22 & 73.72 & 76.49 \\
        {ASAP w/ HNP} &  \textbf{94.38} & \textbf{12.73} & \textbf{87.71} & 88.53 & 81.72 & \textbf{82.89} & \textbf{77.35} & \textbf{84.75} & \textbf{76.54} & \textbf{79.38} & \textbf{73.86} & \textbf{76.52} \\
        \bottomrule[1.5pt]
        \end{tabular}
    }
\caption{\textbf{Discussion of negative patch selection} in Patch Manipulation Modeling. `HNP' is our proposed strategy that only consider the adjacent patches as the negatives.}
\label{tab:patch}
\end{table*}

\noindent\textbf{Full Objectives.} Aligning with prior studies~\cite{HAMMER,VIKI}, we incorporate the DGM\textsuperscript{4} task loss $\mathcal{L}_{\text{DGM}}$, as illustrated in Figure~\ref{framework}, which is composed of four sub-task losses. The bounding box regression loss is determined by the discrepancy between the predicted coordinates of the bounding detector and the annotations. The binary classification loss measures the binary cross-entropy between the classifier's output and the manipulation presence label. The multi-label classifier discerns the manipulation type, with a multi-class classification loss applied to the prediction and annotation. The token detection loss, responsible for identifying manipulated words, implements a binary classification for each word against its label\footnote{We append the comprehensive definition of each loss in our supplementary file.}. Our overall optimization objective is as follows:
\begin{equation}
\label{finaloss}
    \mathcal{L} = \mathcal{L}_{\text{DGM}} + \mathcal{L}_{\text{LMA}} + \alpha\mathcal{L}_{\text{MGCA}} + \lambda\mathcal{L}_{\text{PMM}},
\end{equation}
where $\alpha$ and $\lambda$ are both trade-off hyper-parameters to balance the losses.

\section{Experiments}
\label{exper}
\definecolor{mycolor}{rgb}{0.878, 0.004, 0.831}
\begin{figure*}[h]
    \centering
    \includegraphics[width=1\textwidth]{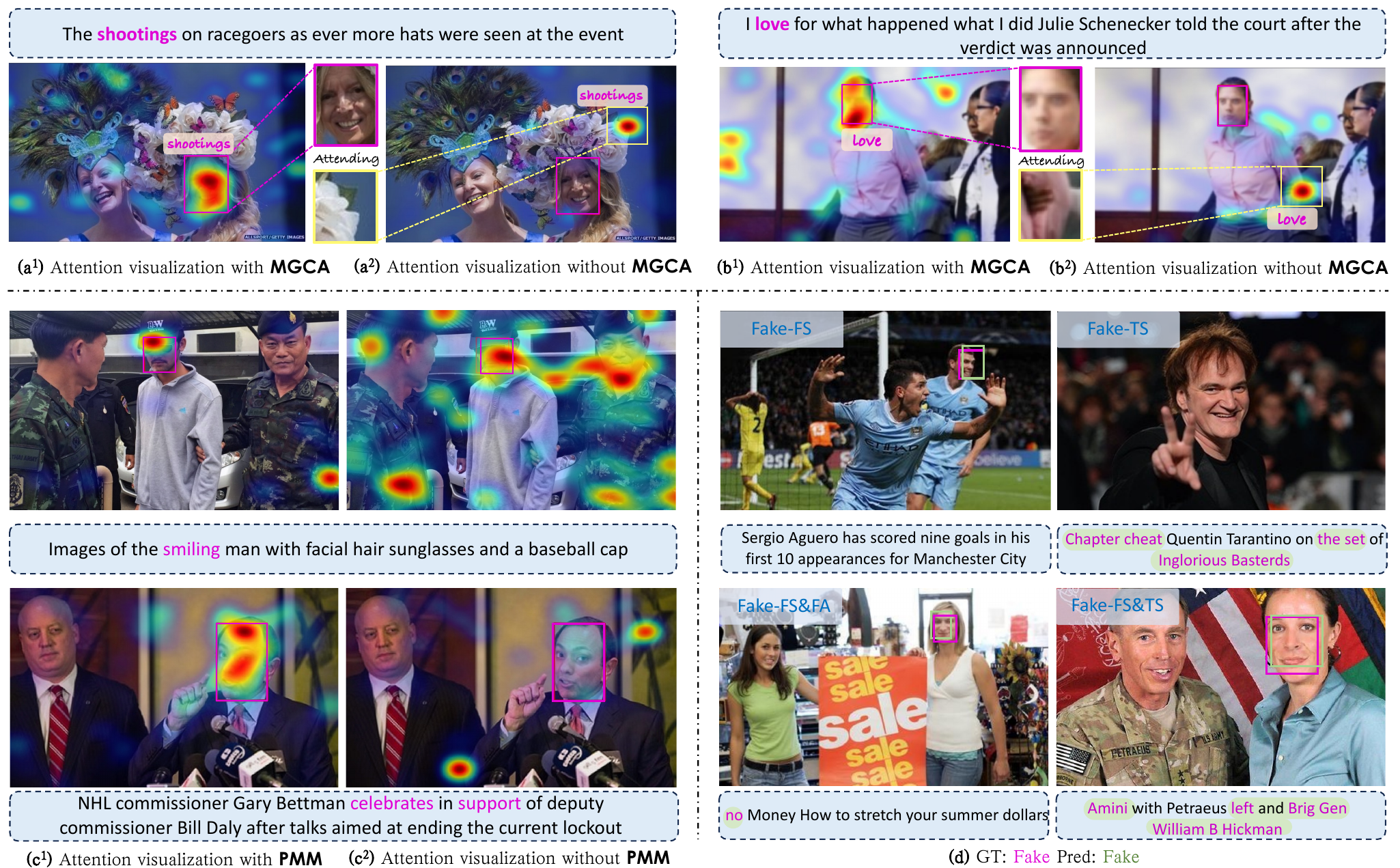}
    \caption{Effect of \textbf{MGCA} and \textbf{PMM} Loss on Attention Map Visualization. The \textcolor{mycolor}{red} rectangle represents the bounding box of the manipulated face, and the \textcolor{mycolor}{red} text indicates the manipulated word. (a) and (b) show the attention visualization between the manipulated word and the image. (c) shows the attention visualization between the entire sentence and the image. (d) presents the model’s prediction compared to the Ground Truth.}
    \label{fig:visual_mpp}
\end{figure*}
\subsection{Implementation Details}

In this experiment, all results are obtained using 8 NVIDIA GeForce RTX 4090 GPUs. For consistency, we resize the images to 256×256. We use the Vision Transformer (ViT-B/16)~\cite{VisionTransformer} as the image encoder and Bert-base~\cite{BERT} as the text encoder.  The pre-trained Multimodal Large Language Model (MLLM) used is VisCPM~\cite{VisCPM}, and the Large Language Model (LLM) is Mistral 7B~\cite{Mistral7B}. MLP consists of three layers, binary classifier $\mathcal{C}$ consists of one layer. In Equations~\ref{detecthead} and~\ref{finaloss}, the learnable parameter $\delta$ is initialized to 0.5, while the parameters are set to $\alpha=0.1$ and $\lambda=0.01$.
Additionally, to ensure fairness, our batch size and training epochs are set to 32 and 50, respectively, consistent with the baseline. During the first 1000 steps, the learning rate is warmed up to $1 \times 10^{-4}$, and then a cosine schedule is used to decay the learning rate to $1 \times 10^{-6}$. The AdamW optimizer is adopted to update the parameters, with a weight decay ratio of 0.02.
\subsection{Dataset and Evaluation Metrics}
We validate the effectiveness of our method on the DGM\textsuperscript{4} dataset~\cite{HAMMER}, which contains 230K image-text pairs. Of these, 33.7\% are original, while the remaining pairs are manipulated, including face swap (FS), face attribute (FA), text swap (TS) and text attribute (TA) manipulation. 
The task is challenging due to mixed manipulated regions and text, but captions and explanations aid the ASAP model.
To assess the performance of ASAP method, We evaluate it on four tasks: manipulated image bounding box grounding, binary classification, multi-classification and manipulated text token grounding. For the manipulated image bounding box grounding task, we use IoUmean, IoU50, and IoU75 as metrics. In the binary classification task, we utilize AUC, EER, and ACC. For the multi-classification task, we assess using mAP, CF1, and OF1. For the manipulated text token grounding task, we measure performance with Precision, Recall, and F1 Score.

\subsection{Quantitative Results}
To evaluate the performance of our contrastive learning approach, we conducted a detailed comparison against state-of-the-art methods. 
Our ASAP method was tested across six popular frameworks using the DGM\textsuperscript{4} dataset and consistently demonstrated superior performance. 
Table~\ref{tab:comparison_HAMMER} highlights the quantitative results, where ASAP consistently outperforms the comparison methods. 
Notably, HAMMER++, VIKI, UFAFormer, and our ASAP all build upon HAMMER as the baseline. DGM\textsuperscript{4}, as a challenging task, especially for the binary classification, significant improvements are hard to observe: HAMMER++ boosts AUC by just 0.24, and even the state-of-the-art UFAFormer manages only a 0.72 increase. However, ASAP achieves a substantial \textbf{1.29} gain in AUC and an average improvement of 2.54 across all tasks. Consequently, we can conclude that our method achieves a notable improvement.
\begin{figure}[t]
    \centering 
    \includegraphics[width=0.4\textwidth]{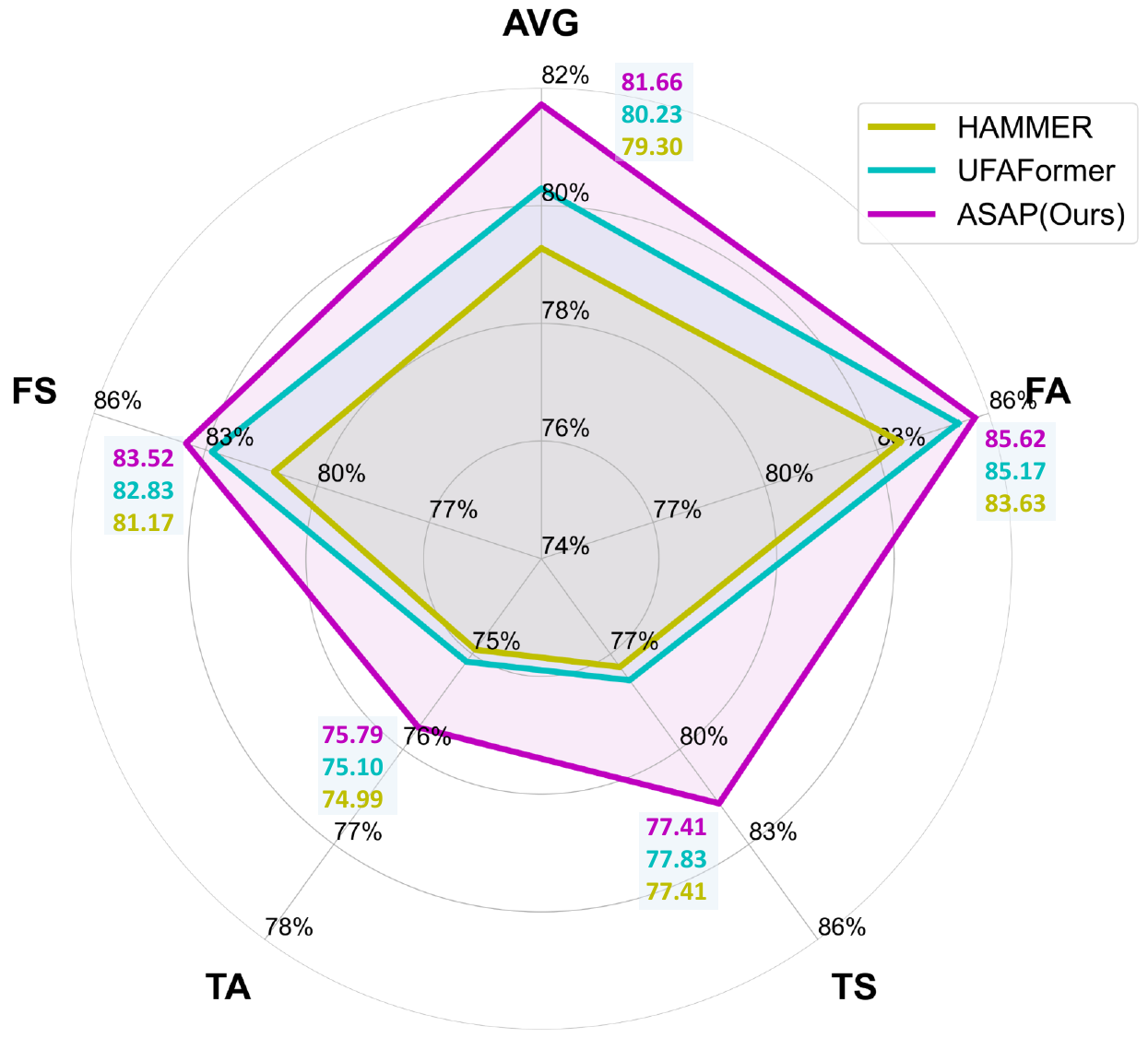} 
    \vspace{-0.3cm}
    \caption{\textbf{F1 score comparison} of HAMMER, UFAFormer and our ASAP. ASAP can surpass HAMMER by a clear margin. } 
    \label{attcomp}
\end{figure}


Notably, AUC and ACC achieve 94.38\% and 87.71\%, surpassing UFAFormer's results.  The multi-label classification mAP reaches 88.53\%, achieving a nearly 2\% improvement over VIKI. For manipulated image grounding, IoUmean is 77.35\%, demonstrating nearly a 1\% improvement over HAMMER and VIKI. In manipulated text token grounding, our method achieves a Precision of 79.38\%, surpassing UFAFormer by 6.03\%. 
ASAP’s performance in text token grounding stands out significantly, with a Precision of 79.38\% and a Recall of 73.86\%, reflecting a 4.3\% precision increase over HAMMER.
Figure~\ref{attcomp} illustrates F1 scores for four manipulation types, with ASAP outperforming HAMMER across all metrics. Specifically, ASAP shows 2\% improvement in Face manipulation and 4.3\% improvement in Text Swap manipulation.
\subsection{Ablation Study}
\noindent\textbf{Component Ablation.} 
Table~\ref{tab:ablation} highlights the contributions of individual modules within our framework. We utilized a retrained HAMMER model as the baseline for comparison. Integrating the Large Model-assisted Alignment (LMA) significantly enhances performance, with AUC exceeding 94\%, marking a 1.12\% improvement, and EER reduced to under 13\%. Text grounding accuracy also improves, reaching 78\%. However, incorporating image features in binary classification results in a relative decline in image grounding accuracy.
The addition of MGCA further boosts performance, yielding a 1.05\% increase in image grounding and a 0.31\% gain in text grounding, surpassing baseline results and ensuring higher overall efficacy. Finally, the inclusion of PMM loss leads to an additional 0.4\% enhancement in image grounding, most of other metrics also achieving optimal performance.

\noindent\textbf{Effectiveness of Caption and Explanation Text.} As demonstrated in Table~\ref{tab:caption}, 
the impact of auxiliary caption and explanation text on performance is evaluated. Utilizing either caption or explanation text individually improves performance over HAMMER, but combining both yields the greatest enhancement, particularly in mAP and IoUmean, which improve by around 0.4\%.

\noindent\textbf{Effectiveness of MGCA and PMM loss.} Figure~\ref{fig:visual_mpp} illustrates the attention visualization after integrating the MGCA and PMM losses. The MGCA loss directs the model to focus on inconsistencies between manipulated facial features and text, enhancing classification accuracy. While the PMM loss strengthens the model’s ability to localize manipulated regions, crucial for the manipulated image grounding task. This demonstrates the distinct roles of MGCA and PMM in improving classification precision and grounding performance, respectively.

\noindent\textbf{Negative Selection in PMM.} To rigorously evaluate the efficacy of our HNP strategy, we conduct a comparison between the ``w/ HNP” and ``w/o HNP” approaches. Table~\ref{tab:patch} reveals that while most metrics remain relatively stable, the HNP strategy enhances the model’s image grounding capability, with the Intersection over Union (IoU) mean increasing by 0.43\%. This demonstrates the effectiveness of selectively considering adjacent patches as negatives.

\section{Conclusion}
\label{conclu}

The paper introduces ASAP, a new framework designed to improve the detection and grounding of multimodal media manipulation (DGM\textsuperscript{4}) by focusing on cross-modal semantic alignment, an aspect often overlooked by existing DGM\textsuperscript{4} approaches. To address this problem, ASAP employs MLLMs and LLMs to create paired image-text instances for manipulated content and applies cross-modal alignment learning for semantic refinement. Furthermore, we develop the MGCA mechanism that provides implicit guidance to help the model better perceive manipulations. During training, MGCA and PMM emphasize manipulated components, improving the model's manipulation-detection capabilities. Experiments demonstrate that ASAP significantly outperforms comparable methods.
\section*{Acknowledgments}

This work was supported by the National Key R\&D Program of China (2024YFB3311602),  the Fundamental Research Funds for the Central Universities, including JZ2024HGTB0261, JZ2023YQTD0072, JZ2024HGTG0309, and JZ2024AHST0337,  the National Natural Science Foundation of China (72188101, 62020106007, 62302140), the Major Project of Anhui Province (202203a05020011), and the Anhui Provincial Natural Science Foundation - Distinguished Young Scholars Program (2408085J040). This work also received support from the New Cornerstone Science Foundation through the XPLORER PRIZE.
{
    \small
    \bibliographystyle{ieeenat_fullname}
    \bibliography{main}

\begin{thebibliography}{30}
\providecommand{\natexlab}[1]{#1}
\providecommand{\url}[1]{\texttt{#1}}
\expandafter\ifx\csname urlstyle\endcsname\relax
  \providecommand{\doi}[1]{doi: #1}\else
  \providecommand{\doi}{doi: \begingroup \urlstyle{rm}\Url}\fi

\bibitem[Abdelnabi et~al.(2022)Abdelnabi, Hasan, and Fritz]{MultilModel}
Sahar Abdelnabi, Rakibul Hasan, and Mario Fritz.
\newblock Open-domain, content-based, multi-modal fact-checking of out-of-context images via online resources.
\newblock In \emph{{IEEE/CVF} Conference on Computer Vision and Pattern Recognition, {CVPR} 2022, New Orleans, LA, USA, June 18-24, 2022}, pages 14920--14929. {IEEE}, 2022.

\bibitem[Anil et~al.(2023)Anil, Borgeaud, Wu, Alayrac, Yu, Soricut, Schalkwyk, Dai, Hauth, Millican, Silver, Petrov, Johnson, Antonoglou, Schrittwieser, Glaese, Chen, Pitler, Lillicrap, Lazaridou, Firat, Molloy, Isard, Barham, Hennigan, Lee, Viola, Reynolds, Xu, Doherty, Collins, Meyer, Rutherford, Moreira, Ayoub, Goel, Tucker, Piqueras, Krikun, Barr, Savinov, Danihelka, Roelofs, White, Andreassen, von Glehn, Yagati, Kazemi, Gonzalez, Khalman, Sygnowski, and et~al.]{Gemini}
Rohan Anil, Sebastian Borgeaud, Yonghui Wu, Jean{-}Baptiste Alayrac, Jiahui Yu, Radu Soricut, Johan Schalkwyk, Andrew~M. Dai, Anja Hauth, Katie Millican, David Silver, Slav Petrov, Melvin Johnson, Ioannis Antonoglou, Julian Schrittwieser, Amelia Glaese, Jilin Chen, Emily Pitler, Timothy~P. Lillicrap, Angeliki Lazaridou, Orhan Firat, James Molloy, Michael Isard, Paul~Ronald Barham, Tom Hennigan, Benjamin Lee, Fabio Viola, Malcolm Reynolds, Yuanzhong Xu, Ryan Doherty, Eli Collins, Clemens Meyer, Eliza Rutherford, Erica Moreira, Kareem Ayoub, Megha Goel, George Tucker, Enrique Piqueras, Maxim Krikun, Iain Barr, Nikolay Savinov, Ivo Danihelka, Becca Roelofs, Ana{\"{\i}}s White, Anders Andreassen, Tamara von Glehn, Lakshman Yagati, Mehran Kazemi, Lucas Gonzalez, Misha Khalman, Jakub Sygnowski, and et al.
\newblock Gemini: {A} family of highly capable multimodal models.
\newblock \emph{CoRR}, abs/2312.11805, 2023.

\bibitem[Devlin et~al.(2019)Devlin, Chang, Lee, and Toutanova]{BERT}
Jacob Devlin, Ming{-}Wei Chang, Kenton Lee, and Kristina Toutanova.
\newblock {BERT:} pre-training of deep bidirectional transformers for language understanding.
\newblock In \emph{Proceedings of the 2019 Conference of the North American Chapter of the Association for Computational Linguistics: Human Language Technologies, {NAACL-HLT} 2019, Minneapolis, MN, USA, June 2-7, 2019, Volume 1 (Long and Short Papers)}, pages 4171--4186. Association for Computational Linguistics, 2019.

\bibitem[Dosovitskiy et~al.(2021{\natexlab{a}})Dosovitskiy, Beyer, Kolesnikov, Weissenborn, Zhai, Unterthiner, Dehghani, Minderer, Heigold, Gelly, Uszkoreit, and Houlsby]{VisionTransformer}
Alexey Dosovitskiy, Lucas Beyer, Alexander Kolesnikov, Dirk Weissenborn, Xiaohua Zhai, Thomas Unterthiner, Mostafa Dehghani, Matthias Minderer, Georg Heigold, Sylvain Gelly, Jakob Uszkoreit, and Neil Houlsby.
\newblock An image is worth 16x16 words: Transformers for image recognition at scale.
\newblock In \emph{9th International Conference on Learning Representations, {ICLR} 2021, Virtual Event, Austria, May 3-7, 2021}. OpenReview.net, 2021{\natexlab{a}}.

\bibitem[Dosovitskiy et~al.(2021{\natexlab{b}})Dosovitskiy, Beyer, Kolesnikov, Weissenborn, Zhai, Unterthiner, Dehghani, Minderer, Heigold, Gelly, Uszkoreit, and Houlsby]{vit}
Alexey Dosovitskiy, Lucas Beyer, Alexander Kolesnikov, Dirk Weissenborn, Xiaohua Zhai, Thomas Unterthiner, Mostafa Dehghani, Matthias Minderer, Georg Heigold, Sylvain Gelly, Jakob Uszkoreit, and Neil Houlsby.
\newblock An image is worth 16x16 words: Transformers for image recognition at scale.
\newblock In \emph{9th International Conference on Learning Representations, {ICLR} 2021, Virtual Event, Austria, May 3-7, 2021}. OpenReview.net, 2021{\natexlab{b}}.

\bibitem[Hu et~al.(2024)Hu, Yao, Wang, Wang, Pan, Chen, Yu, Wu, Zhao, Zhang, Han, Lin, Xue, Li, Liu, and Sun]{VisCPM}
Jinyi Hu, Yuan Yao, Chongyi Wang, Shan Wang, Yinxu Pan, Qianyu Chen, Tianyu Yu, Hanghao Wu, Yue Zhao, Haoye Zhang, Xu Han, Yankai Lin, Jiao Xue, Dahai Li, Zhiyuan Liu, and Maosong Sun.
\newblock Large multilingual models pivot zero-shot multimodal learning across languages.
\newblock In \emph{The Twelfth International Conference on Learning Representations, {ICLR} 2024, Vienna, Austria, May 7-11, 2024}. OpenReview.net, 2024.

\bibitem[Jiang et~al.(2023)Jiang, Sablayrolles, Mensch, Bamford, Chaplot, de~Las~Casas, Bressand, Lengyel, Lample, Saulnier, Lavaud, Lachaux, Stock, Scao, Lavril, Wang, Lacroix, and Sayed]{Mistral7B}
Albert~Q. Jiang, Alexandre Sablayrolles, Arthur Mensch, Chris Bamford, Devendra~Singh Chaplot, Diego de Las~Casas, Florian Bressand, Gianna Lengyel, Guillaume Lample, Lucile Saulnier, L{\'{e}}lio~Renard Lavaud, Marie{-}Anne Lachaux, Pierre Stock, Teven~Le Scao, Thibaut Lavril, Thomas Wang, Timoth{\'{e}}e Lacroix, and William~El Sayed.
\newblock Mistral 7b.
\newblock \emph{CoRR}, abs/2310.06825, 2023.

\bibitem[Kim et~al.(2021)Kim, Son, and Kim]{VILT}
Wonjae Kim, Bokyung Son, and Ildoo Kim.
\newblock Vilt: Vision-and-language transformer without convolution or region supervision.
\newblock In \emph{Proceedings of the 38th International Conference on Machine Learning, {ICML} 2021, 18-24 July 2021, Virtual Event}, pages 5583--5594. {PMLR}, 2021.

\bibitem[Li et~al.(2023)Li, Li, Savarese, and Hoi]{Blip2}
Junnan Li, Dongxu Li, Silvio Savarese, and Steven C.~H. Hoi.
\newblock {BLIP-2:} bootstrapping language-image pre-training with frozen image encoders and large language models.
\newblock In \emph{International Conference on Machine Learning, {ICML} 2023, 23-29 July 2023, Honolulu, Hawaii, {USA}}, pages 19730--19742. {PMLR}, 2023.

\bibitem[Li et~al.(2024{\natexlab{a}})Li, Tian, Hu, Luo, and Ma]{MMCode}
Kaixin Li, Yuchen Tian, Qisheng Hu, Ziyang Luo, and Jing Ma.
\newblock Mmcode: Evaluating multi-modal code large language models with visually rich programming problems.
\newblock \emph{CoRR}, abs/2404.09486, 2024{\natexlab{a}}.

\bibitem[Li et~al.(2024{\natexlab{b}})Li, Gao, Zhang, Zhai, Chen, and Jeon]{VIKI}
Qilei Li, Mingliang Gao, Guisheng Zhang, Wenzhe Zhai, Jinyong Chen, and Gwanggil Jeon.
\newblock Towards multimodal disinformation detection by vision-language knowledge interaction.
\newblock \emph{Inf. Fusion}, 102:\penalty0 102037, 2024{\natexlab{b}}.

\bibitem[Lin et~al.(2024)Lin, He, Ju, Wang, Ding, and Hu]{deepfake}
Li Lin, Xinan He, Yan Ju, Xin Wang, Feng Ding, and Shu Hu.
\newblock Preserving fairness generalization in deepfake detection.
\newblock In \emph{Proceedings of the IEEE/CVF Conference on Computer Vision and Pattern Recognition}, pages 16815--16825, 2024.

\bibitem[Liu et~al.(2023{\natexlab{a}})Liu, Tan, Chen, Wei, Zhao, and Wang]{UFAFormer}
Huan Liu, Zichang Tan, Qiang Chen, Yunchao Wei, Yao Zhao, and Jingdong Wang.
\newblock Unified frequency-assisted transformer framework for detecting and grounding multi-modal manipulation.
\newblock \emph{CoRR}, abs/2309.09667, 2023{\natexlab{a}}.

\bibitem[Liu et~al.(2023{\natexlab{b}})Liu, Tan, Chen, Wei, Zhao, and Wang]{UFTFormer}
Huan Liu, Zichang Tan, Qiang Chen, Yunchao Wei, Yao Zhao, and Jingdong Wang.
\newblock Unified frequency-assisted transformer framework for detecting and grounding multi-modal manipulation.
\newblock \emph{CoRR}, abs/2309.09667, 2023{\natexlab{b}}.

\bibitem[Luo et~al.(2021)Luo, Darrell, and Rohrbach]{NewsCLIP}
Grace Luo, Trevor Darrell, and Anna Rohrbach.
\newblock Newsclippings: Automatic generation of out-of-context multimodal media.
\newblock In \emph{Proceedings of the 2021 Conference on Empirical Methods in Natural Language Processing, {EMNLP} 2021, Virtual Event / Punta Cana, Dominican Republic, 7-11 November, 2021}, pages 6801--6817. Association for Computational Linguistics, 2021.

\bibitem[OpenAI(2023)]{GPT4}
OpenAI.
\newblock {GPT-4} technical report.
\newblock \emph{CoRR}, abs/2303.08774, 2023.

\bibitem[Peebles and Xie(2023)]{DIT}
William Peebles and Saining Xie.
\newblock Scalable diffusion models with transformers.
\newblock In \emph{{IEEE/CVF} International Conference on Computer Vision, {ICCV} 2023, Paris, France, October 1-6, 2023}, pages 4172--4182. {IEEE}, 2023.

\bibitem[Radford et~al.(2021)Radford, Kim, Hallacy, Ramesh, Goh, Agarwal, Sastry, Askell, Mishkin, Clark, Krueger, and Sutskever]{CLIP}
Alec Radford, Jong~Wook Kim, Chris Hallacy, Aditya Ramesh, Gabriel Goh, Sandhini Agarwal, Girish Sastry, Amanda Askell, Pamela Mishkin, Jack Clark, Gretchen Krueger, and Ilya Sutskever.
\newblock Learning transferable visual models from natural language supervision.
\newblock In \emph{Proceedings of the 38th International Conference on Machine Learning, {ICML} 2021, 18-24 July 2021, Virtual Event}, pages 8748--8763. {PMLR}, 2021.

\bibitem[Reed et~al.(2022)Reed, Zolna, Parisotto, Colmenarejo, Novikov, Barth{-}Maron, Gimenez, Sulsky, Kay, Springenberg, Eccles, Bruce, Razavi, Edwards, Heess, Chen, Hadsell, Vinyals, Bordbar, and de~Freitas]{Gato}
Scott~E. Reed, Konrad Zolna, Emilio Parisotto, Sergio~Gomez Colmenarejo, Alexander Novikov, Gabriel Barth{-}Maron, Mai Gimenez, Yury Sulsky, Jackie Kay, Jost~Tobias Springenberg, Tom Eccles, Jake Bruce, Ali Razavi, Ashley Edwards, Nicolas Heess, Yutian Chen, Raia Hadsell, Oriol Vinyals, Mahyar Bordbar, and Nando de Freitas.
\newblock A generalist agent.
\newblock \emph{CoRR}, abs/2205.06175, 2022.

\bibitem[Rombach et~al.(2021)Rombach, Blattmann, Lorenz, Esser, and Ommer]{diffusion}
Robin Rombach, Andreas Blattmann, Dominik Lorenz, Patrick Esser, and Bj{\"{o}}rn Ommer.
\newblock High-resolution image synthesis with latent diffusion models.
\newblock \emph{CoRR}, abs/2112.10752, 2021.

\bibitem[Shao et~al.(2023)Shao, Wu, and Liu]{HAMMER}
Rui Shao, Tianxing Wu, and Ziwei Liu.
\newblock Detecting and grounding multi-modal media manipulation.
\newblock In \emph{{IEEE/CVF} Conference on Computer Vision and Pattern Recognition, {CVPR} 2023, Vancouver, BC, Canada, June 17-24, 2023}, pages 6904--6913. {IEEE}, 2023.

\bibitem[Shao et~al.(2024)Shao, Wu, Wu, Nie, and Liu]{HAMMER++}
Rui Shao, Tianxing Wu, Jianlong Wu, Liqiang Nie, and Ziwei Liu.
\newblock Detecting and grounding multi-modal media manipulation and beyond.
\newblock \emph{{IEEE} Trans. Pattern Anal. Mach. Intell.}, 46\penalty0 (8):\penalty0 5556--5574, 2024.

\bibitem[Teng et~al.(2024)Teng, Zheng, Ding, Hong, Wangni, Yang, and Tang]{relay_diffusion}
Jiayan Teng, Wendi Zheng, Ming Ding, Wenyi Hong, Jianqiao Wangni, Zhuoyi Yang, and Jie Tang.
\newblock Relay diffusion: Unifying diffusion process across resolutions for image synthesis.
\newblock In \emph{The Twelfth International Conference on Learning Representations, {ICLR} 2024, Vienna, Austria, May 7-11, 2024}. OpenReview.net, 2024.

\bibitem[Touvron et~al.(2023)Touvron, Lavril, Izacard, Martinet, Lachaux, Lacroix, Rozi{\`{e}}re, Goyal, Hambro, Azhar, Rodriguez, Joulin, Grave, and Lample]{llama}
Hugo Touvron, Thibaut Lavril, Gautier Izacard, Xavier Martinet, Marie{-}Anne Lachaux, Timoth{\'{e}}e Lacroix, Baptiste Rozi{\`{e}}re, Naman Goyal, Eric Hambro, Faisal Azhar, Aur{\'{e}}lien Rodriguez, Armand Joulin, Edouard Grave, and Guillaume Lample.
\newblock Llama: Open and efficient foundation language models.
\newblock \emph{CoRR}, abs/2302.13971, 2023.

\bibitem[Vaswani et~al.(2017)Vaswani, Shazeer, Parmar, Uszkoreit, Jones, Gomez, Kaiser, and Polosukhin]{transformer}
Ashish Vaswani, Noam Shazeer, Niki Parmar, Jakob Uszkoreit, Llion Jones, Aidan~N. Gomez, Lukasz Kaiser, and Illia Polosukhin.
\newblock Attention is all you need.
\newblock In \emph{NeurIPS}, pages 5998--6008, 2017.

\bibitem[Wang et~al.(2024)Wang, Bai, Tan, Wang, Fan, Bai, Chen, Liu, Wang, Ge, Fan, Dang, Du, Ren, Men, Liu, Zhou, Zhou, and Lin]{QWEN}
Peng Wang, Shuai Bai, Sinan Tan, Shijie Wang, Zhihao Fan, Jinze Bai, Keqin Chen, Xuejing Liu, Jialin Wang, Wenbin Ge, Yang Fan, Kai Dang, Mengfei Du, Xuancheng Ren, Rui Men, Dayiheng Liu, Chang Zhou, Jingren Zhou, and Junyang Lin.
\newblock Qwen2-vl: Enhancing vision-language model's perception of the world at any resolution.
\newblock \emph{CoRR}, abs/2409.12191, 2024.

\bibitem[Wang et~al.(2018)Wang, Ma, Jin, Yuan, Xun, Jha, Su, and Gao]{MultilModel2}
Yaqing Wang, Fenglong Ma, Zhiwei Jin, Ye Yuan, Guangxu Xun, Kishlay Jha, Lu Su, and Jing Gao.
\newblock {EANN:} event adversarial neural networks for multi-modal fake news detection.
\newblock In \emph{Proceedings of the 24th {ACM} {SIGKDD} International Conference on Knowledge Discovery {\&} Data Mining, {KDD} 2018, London, UK, August 19-23, 2018}, pages 849--857. {ACM}, 2018.

\bibitem[Xuan et~al.(2019)Xuan, Peng, Wang, and Dong]{Unimodel1Xue}
Xinsheng Xuan, Bo Peng, Wei Wang, and Jing Dong.
\newblock On the generalization of {GAN} image forensics.
\newblock In \emph{Biometric Recognition - 14th Chinese Conference, {CCBR} 2019, Zhuzhou, China, October 12-13, 2019, Proceedings}, pages 134--141. Springer, 2019.

\bibitem[Xue et~al.(2023)Xue, Jiang, Liu, and Wei]{Unimodel2}
Ziyu Xue, Xiuhua Jiang, Qingtong Liu, and Zhaoshan Wei.
\newblock Global-local facial fusion based {GAN} generated fake face detection.
\newblock \emph{Sensors}, 23\penalty0 (2):\penalty0 616, 2023.

\bibitem[Zeng et~al.(2024)Zeng, Xu, Wang, Zhang, Yin, Rojas, Feng, Zhao, Lai, Yu, Wang, Sun, Zhang, Cheng, Gui, Tang, Zhang, Li, Zhao, Wu, Zhong, Liu, Huang, Zhang, Zheng, Lu, Duan, Zhang, Cao, Yang, Tam, Zhao, Liu, Xia, Zhang, Gu, Lv, Liu, Liu, Yang, Song, Zhang, An, Xu, Niu, Yang, Li, Bai, Dong, Qi, Wang, Yang, Du, Hou, and Wang]{GLM}
Aohan Zeng, Bin Xu, Bowen Wang, Chenhui Zhang, Da Yin, Diego Rojas, Guanyu Feng, Hanlin Zhao, Hanyu Lai, Hao Yu, Hongning Wang, Jiadai Sun, Jiajie Zhang, Jiale Cheng, Jiayi Gui, Jie Tang, Jing Zhang, Juanzi Li, Lei Zhao, Lindong Wu, Lucen Zhong, Mingdao Liu, Minlie Huang, Peng Zhang, Qinkai Zheng, Rui Lu, Shuaiqi Duan, Shudan Zhang, Shulin Cao, Shuxun Yang, Weng~Lam Tam, Wenyi Zhao, Xiao Liu, Xiao Xia, Xiaohan Zhang, Xiaotao Gu, Xin Lv, Xinghan Liu, Xinyi Liu, Xinyue Yang, Xixuan Song, Xunkai Zhang, Yifan An, Yifan Xu, Yilin Niu, Yuantao Yang, Yueyan Li, Yushi Bai, Yuxiao Dong, Zehan Qi, Zhaoyu Wang, Zhen Yang, Zhengxiao Du, Zhenyu Hou, and Zihan Wang.
\newblock Chatglm: {A} family of large language models from {GLM-130B} to {GLM-4} all tools.
\newblock \emph{CoRR}, abs/2406.12793, 2024.

\end{thebibliography}
}

\clearpage
\setcounter{page}{1}
\maketitlesupplementary

\section{DGM\textsuperscript{4} Loss}
Given an image-text pair $(I, T)$, we define four sub-task losses following HAMMER as follows:
\subsection{Manipulated Image Bounding Box Grounding}
For the manipulated image grounding task, we input the multimodal feature $i_{pat}^v$ into a BBox Detector $D_v$ and calculate the Image Manipulation Grounding Loss as:
\begin{equation}
\begin{aligned}
\mathcal{L}_{\text{IMG}} = \mathbb{E}_{(I,T)} \left[ \| \text{Sigmoid}(D_v(i_{pat}^v)) - y_{box} \| \right. \nonumber \\
\left. + \mathcal{L}_{IOU}(\text{Sigmoid}(D_v(i_{pat}^v)), y_{box}) \right]
\end{aligned}
\end{equation}

\subsection{Binary Classification}
For the binary classification task, we input multimodal feature $M_{it}$ into Binary Classifier $C_b$ and calculate Binary Classifier Loss as follows:
\begin{equation}
\left\{\begin{aligned}
    &\mathcal{L}_{\text{IMG}} = \mathbb{E}_{(I,T)} \left[\boldsymbol{H}(C_b(M_{it}), y_{bin}) \right]\\
    &M_{it}={\delta}\mathcal{E}_{mm}^v(\mathcal{E}_v(I),\mathcal{E}_t(T)) + \mathcal{E}_{mm}^t(\mathcal{E}_t(T),\mathcal{E}_v(I)) \nonumber
\end{aligned}
\right.
\end{equation}
where $\boldsymbol{H}(\cdot)$ is the cross-entropy function.
\subsection{Manipulation Type Detection}
For detection of the type of manipulation task, we input the multimodal feature $M_{it}$ into the Binary Classifier $C_m$ and compute the loss of the binary classifier as:

\begin{equation}
\begin{aligned}
    \mathcal{L}_{\text{MLC}} = \mathbb{E}_{(I,T)} \left[\boldsymbol{H}(C_m(M_{it}), y_{mul}) \right] \nonumber
\end{aligned}
\end{equation}
\subsection{Manipulated Text Token Grounding}
For the manipulated text token grounding task, we use a Token Detector $D_t$ to predict the label of each token in $t_{tok}^t$ and calculate the cross-entropy loss as follows:
\begin{equation}
    \left\{\begin{aligned}
        &\mathcal{L}_{\text{TMG}}=(1-\alpha)\mathcal{L}_{\text{tok}} + \alpha\mathcal{L}_{tok}^{mom}\\
        &\mathcal{L}_{\text{tok}}=\mathbb{E}_{(I,T)} \left[\boldsymbol{H}(D_t(t_{tok}^t), y_{tok}) \right]\\
        &\mathcal{L}_{tok}^{mom}=\mathbb{E}_{(I,T)} KL\left[D_t(t_{tok}^t)|| \hat{D_t}(\hat{t}_{tok}^t)\right]\\
        &\{t_{cls}^t , t_{tok}^t\}=\mathcal{E}_{mm}^t(\mathcal{E}_t(T),\mathcal{E}_v(I)) \nonumber
    \end{aligned}
    \right.
\end{equation}
where $\hat{D_t}(\hat{t}_{tok}^t)$ represents the pseudo-labels generated by the momentum Token Detector, used to modulate the original token predictions, and $\text{KL}$ denotes the Kullback-Leibler divergence between the original token predictions and the momentum-based pseudo-labels.

\section{Discussion}
\subsection{Effectiveness of Cross-modality learning}
We evaluated the multimodal fusion mechanism by comparing single-modal and multimodal learning. Table~\ref{tab:text} and Table~\ref{tab:image} show that “Text” and “Image” represent single-modal learning, while “MultiModal” indicates multimodal learning. The results confirm that our ASAP model improves detection and grounding through multimodal fusion. The difference between the two "MultiModal" results stems from the use of different loss functions.

\begin{table}[t]
    \centering
    \setlength{\tabcolsep}{1.1mm}
    \setlength{\extrarowheight}{0pt}{
    \begin{tabular}{lcccccc}
    \toprule[1.5pt]
         \rowcolor[gray]{.95}Tasks & \multicolumn{3}{c}{Binary Cls} &\multicolumn{3}{c} {Text Grounding}\\
         \cmidrule(r){2-4} \cmidrule(r){5-7}
         \rowcolor[gray]{.95}Methods  & AUC$\uparrow$ & EER$\downarrow$  & ACC$\uparrow$ & mAP$\uparrow$ & CF1$\uparrow$ & OF1$\uparrow$\\
         \midrule
         {Text} & 92.89 & 13.26 & 86.43 & 78.90 & 72.98 & 75.01\\
         {MultiModal} & \textbf{94.58} & \textbf{12.79} & \textbf{87.64}  & \textbf{79.79} & \textbf{73.40} & \textbf{76.46}\\
    \bottomrule[1.5pt]
    \end{tabular}
    }
    \caption{Ablation study of text modality.}
    \label{tab:text}
\end{table}
\subsection{Discussion of different Large Models} 
To assess the effectiveness of our approach and the validity of large model selection, we employed Qwen and LLaMA 2b in the LMA mechanism, conducting ablation studies against our ASAP method. Table~\ref{tab:model} demonstrate that incorporating preliminary texts significantly improves performance across all tasks except image grounding, confirming our approach as the optimal solution.
\subsection{Discussion of each Hyperparameter }
We fine-tuned multiple hyperparameters for the ASAP model, selecting final values of $\delta=0.5$, $\alpha=0.1$, and $\lambda=0.01$ based on model performance. As shown in Tables~\ref{tab:delta},\ref{tab:alpha}, and\ref{tab:lambda}, these values provided a balanced optimization of various performance metrics, enabling ASAP to achieve peak results while preserving efficient detection capabilities.

\begin{table}[t]
    \centering
    \setlength{\tabcolsep}{1.1mm}
    \setlength{\extrarowheight}{0pt}{
    \begin{tabular}{lcccccc}
    \toprule[1.5pt]
         \rowcolor[gray]{.95}Tasks & \multicolumn{3}{c}{Binary Cls} &\multicolumn{3}{c} {Image Grounding}\\
         \cmidrule(r){2-4} \cmidrule(r){5-7}
         \rowcolor[gray]{.95}Methods  & AUC$\uparrow$ & EER$\downarrow$  & ACC$\uparrow$ & mAP$\uparrow$ & CF1$\uparrow$ & OF1$\uparrow$\\
         \midrule
         {Image} & 93.13 & 13.48 & 86.55 & 76.16 & 83.46 & 75.13\\
         {MultiModal} & \textbf{94.50} & \textbf{12.62} & \textbf{87.27} & \textbf{77.30} & \textbf{84.22} & \textbf{77.61} \\
    \bottomrule[1.5pt]
    \end{tabular}
    }
    \caption{Ablation study of image modality.}
    \label{tab:image}
\end{table}

\begin{table*}[h]
\centering
\small
\setlength{\tabcolsep}{0.8mm}
\setlength{\extrarowheight}{0pt}{
    \begin{tabular}{lcccccccccccc}
    \toprule[1.5pt]
    \rowcolor[gray]{.95}Tasks & \multicolumn{3}{c}{Binary Cls} &\multicolumn{3}{c} {Multi-Label} &  \multicolumn{3}{c}{Image Grounding} & \multicolumn{3}{c}{Text Grounding} \\
    \cmidrule(r){2-4} \cmidrule(r){5-7}\cmidrule(r){8-10}\cmidrule(r){11-13}
    \rowcolor[gray]{.95}Different models  & AUC$\uparrow$ & EER$\downarrow$  & ACC$\uparrow$ & mAP$\uparrow$ & CF1$\uparrow$ & OF1$\uparrow$ & IoUmean$\uparrow$ & IoU50$\uparrow$ & IoU75$\uparrow$ & Precision$\uparrow$ & Recall$\uparrow$ & F1$\uparrow$ \\
    \midrule
    Baseline & 93.16 & 14.13 & 86.23 & 86.23 & 79.59 & 80.54 & \textbf{76.49} & \textbf{83.82} & \textbf{75.97} & 75.25 & 68.21 & 71.83 \\
    Qwen-VL \& LLaMA & 94.23 & \textbf{12.83} & 87.49 & 86.90 & 80.15 & 82.01 & 75.78 & 83.21 & 74.86 & 78.44 & 73.60 & 75.04 \\
    VisCPM \& Mistral & \textbf{94.28} &12.86 & \textbf{87.53} & \textbf{88.10} & \textbf{81.71} & \textbf{82.61} & 75.90 & 83.27 & 74.98 & \textbf{78.59} & \textbf{74.10} & \textbf{76.28} \\
    \bottomrule[1.5pt]
    \end{tabular}
}
\caption{Performance Comparison Across Different Large Models of \textbf{LMA}. This table compares two approaches of large model assistance, where Mistral is used as the LLM and Viscpm as the MLLM, and LLaMA is used as the LLM and Qwen VL as the MLLM, to generate auxiliary labels in the LMA module.}
\label{tab:model}
\end{table*}

\begin{table*}[h]
\centering
\small
\setlength{\tabcolsep}{0.8mm}
\setlength{\extrarowheight}{0pt}{
    \begin{tabular}{lcccccccccccc}
    \toprule[1.5pt]
    \rowcolor[gray]{.95}Tasks & \multicolumn{3}{c}{Binary Cls} &\multicolumn{3}{c} {Multi-Label} &  \multicolumn{3}{c}{Image Grounding} & \multicolumn{3}{c}{Text Grounding} \\
    \cmidrule(r){2-4} \cmidrule(r){5-7}\cmidrule(r){8-10}\cmidrule(r){11-13}
    \rowcolor[gray]{.95}Different $\delta$  & AUC$\uparrow$ & EER$\downarrow$  & ACC$\uparrow$ & mAP$\uparrow$ & CF1$\uparrow$ & OF1$\uparrow$ & IoUmean$\uparrow$ & IoU50$\uparrow$ & IoU75$\uparrow$ & Precision$\uparrow$ & Recall$\uparrow$ & F1$\uparrow$ \\
    \midrule
    \ $\delta =0.1$ & 94.33 & 12.89 & 87.75 & 88.32 & 81.57 & 82.66 & 77.01 & 84.23 & 76.19 & 79.02 & 73.72 & \textbf{76.66} \\
    \ $\delta =0.5$ & \textbf{94.38} & \textbf{12.73} & 87.71 & \textbf{88.53} & \textbf{81.72} & \textbf{82.89} & \textbf{77.35} & \textbf{84.75} & \textbf{76.54} & \textbf{79.38} & \textbf{73.86} & 76.52 \\
    \ $\delta =1.0$ & 94.32 & 12.79 & \textbf{87.77} & 88.35 & 81.71 & 82.88 & 77.03 & 84.45 & 76.10 & 78.66 & 73.71 & 76.11 \\
    \bottomrule[1.5pt]
    \end{tabular}
}
\caption{Performance Comparison Across Different Initial Values of the \textbf{Hyperparameter $\delta$} in equation~\ref{detecthead}.}
\label{tab:delta}
\end{table*}

\begin{table*}[h]
\centering
\small
\setlength{\tabcolsep}{0.8mm}
\setlength{\extrarowheight}{0pt}{
    \begin{tabular}{lcccccccccccc}
    \toprule[1.5pt]
    \rowcolor[gray]{.95}Tasks & \multicolumn{3}{c}{Binary Cls} &\multicolumn{3}{c} {Multi-Label} &  \multicolumn{3}{c}{Image Grounding} & \multicolumn{3}{c}{Text Grounding} \\
    \cmidrule(r){2-4} \cmidrule(r){5-7}\cmidrule(r){8-10}\cmidrule(r){11-13}
    \rowcolor[gray]{.95}Different $\alpha$  & AUC$\uparrow$ & EER$\downarrow$  & ACC$\uparrow$ & mAP$\uparrow$ & CF1$\uparrow$ & OF1$\uparrow$ & IoUmean$\uparrow$ & IoU50$\uparrow$ & IoU75$\uparrow$ & Precision$\uparrow$ & Recall$\uparrow$ & F1$\uparrow$ \\
    \midrule
    \ $\alpha =0.1$ & \textbf{94.38} & \textbf{12.73} & \textbf{87.71} & \textbf{88.53} & \textbf{81.72} & \textbf{82.89} & 77.35 & 84.75 & 76.54 & \textbf{79.38} & \textbf{73.86} & \textbf{76.52} \\
    \ $\alpha =0.5$ & 94.30 & 12.60 & 87.65 & 88.05 & 81.71 & 82.81 & \textbf{77.51} & \textbf{84.83} & \textbf{77.12} & 79.31 & 72.79 & 75.92\\
    \ $\alpha =1.0$ & 94.26 & 12.99 & 87.32 & 87.98 & 81.55 & 82.34 & 77.23 & 84.48 & 76.57 & 78.90 & 72.45 & 75.68\\
    \bottomrule[1.5pt]
    \end{tabular}
}
\caption{Performance Comparison Across Different Values of the \textbf{Hyperparameter $\alpha$} in equation~\ref{finaloss}.}
\label{tab:alpha}
\end{table*}

\begin{table*}[h]
\centering
\small
\setlength{\tabcolsep}{0.8mm}
\setlength{\extrarowheight}{0pt}{
    \begin{tabular}{lcccccccccccc}
    \toprule[1.5pt]
    \rowcolor[gray]{.95}Tasks & \multicolumn{3}{c}{Binary Cls} &\multicolumn{3}{c} {Multi-Label} &  \multicolumn{3}{c}{Image Grounding} & \multicolumn{3}{c}{Text Grounding} \\
    \cmidrule(r){2-4} \cmidrule(r){5-7}\cmidrule(r){8-10}\cmidrule(r){11-13}
    \rowcolor[gray]{.95}Different $\lambda$  & AUC$\uparrow$ & EER$\downarrow$  & ACC$\uparrow$ & mAP$\uparrow$ & CF1$\uparrow$ & OF1$\uparrow$ & IoUmean$\uparrow$ & IoU50$\uparrow$ & IoU75$\uparrow$ & Precision$\uparrow$ & Recall$\uparrow$ & F1$\uparrow$ \\
    \midrule
    \ $\lambda =0.01$ & 94.38 & \textbf{12.73} & \textbf{87.71} & 88.53 & \textbf{81.72} & \textbf{82.89} & \textbf{77.35} & \textbf{84.75} & \textbf{76.54} & 79.38 & \textbf{73.86} & \textbf{76.52} \\
    \ $\lambda =0.05$ & \textbf{94.42} & 12.80 & 87.63 & \textbf{88.59} & 81.68 & 82.79 & 77.19 & 84.49 & 76.61 & \textbf{79.44} & 73.85 & 76.49 \\
    \ $\lambda =0.10$ &  94.35 & 12.86 & 87.55 & 88.55 & 81.71 & 82.80 & 77.10 & 84.38 & 76.52 & 79.21 & 73.84 & 75.98\\
    \bottomrule[1.5pt]
    \end{tabular}
}
\caption{Performance Comparison Across Different Values of the \textbf{Hyperparameter $\lambda$} in equation~\ref{finaloss}.}
\label{tab:lambda}
\end{table*}

\end{document}